\documentclass[a4paper,conference]{IEEEtran}

\usepackage{cite}
\usepackage[pdftex]{graphicx}
\usepackage{amsmath}
\usepackage{amsfonts}
\usepackage{bbm}
\usepackage{dsfont}
\usepackage[ruled,vlined,linesnumbered]{algorithm2e}
\usepackage{cleveref}
\usepackage{algorithmic}
\usepackage{array}
\usepackage[caption=false,font=footnotesize]{subfig}
\usepackage{fixltx2e}
\usepackage{dblfloatfix}
\usepackage{url}
\usepackage{amsfonts}

\begin{document}
\title{Real-Time Accident Detection in Traffic Surveillance Using Deep Learning}

\author{
	\IEEEauthorblockN{Hadi Ghahremannezhad}
	\IEEEauthorblockA{Department of Computer Science\\
		New Jersey Institute of Technology\\
		Newark, NJ 07102, USA\\
		Email: hg255@njit.edu}
	\and
	\IEEEauthorblockN{Hang Shi}
	\IEEEauthorblockA{Innovative AI Technologies\\
		Newark, NJ 07103, USA\\
		Email: hang@iaitusa.com}
	\and
	\IEEEauthorblockN{Chengjun Liu}
	\IEEEauthorblockA{Department of Computer Science\\
		New Jersey Institute of Technology\\
		Newark, NJ 07102, USA\\
		Email: cliu@njit.edu}
}

\maketitle

\begin{abstract}
	Automatic detection of traffic accidents is an important emerging topic in traffic monitoring systems. Nowadays many urban intersections are equipped with surveillance cameras connected to traffic management systems. Therefore, computer vision techniques can be viable tools for automatic accident detection. This paper presents a new efficient framework for accident detection at intersections for traffic surveillance applications. The proposed framework consists of three hierarchical steps, including efficient and accurate object detection based on the state-of-the-art YOLOv4 method, object tracking based on Kalman filter coupled with the Hungarian algorithm for association, and accident detection by trajectory conflict analysis. A new cost function is applied for object association to accommodate for occlusion, overlapping objects, and shape changes in the object tracking step. The object trajectories are analyzed in terms of velocity, angle, and distance in order to detect different types of trajectory conflicts including vehicle-to-vehicle, vehicle-to-pedestrian, and vehicle-to-bicycle. Experimental results using real traffic video data show the feasibility of the proposed method in real-time applications of traffic surveillance. In particular, trajectory conflicts, including near-accidents and accidents occurring at urban intersections are detected with a low false alarm rate and a high detection rate. The robustness of the proposed framework is evaluated using video sequences collected from YouTube with diverse illumination conditions. The dataset is publicly available at: http://github.com/hadi-ghnd/AccidentDetection.
\end{abstract}


\section{Introduction}\label{sec_intr}
One of the main problems in urban traffic management is the conflicts and accidents occurring at the intersections.
Drivers caught in a dilemma zone may decide to accelerate at the time of phase change from green to yellow, which in turn may induce rear-end and angle crashes. 
Additionally, despite all the efforts in preventing hazardous driving behaviors, running the red light is still common.
Other dangerous behaviors, such as sudden lane changing and unpredictable pedestrian/cyclist movements at the intersection, may also arise due to the nature of traffic control systems or intersection geometry.
Timely detection of such trajectory conflicts is necessary for devising countermeasures to mitigate their potential harms.

Currently, most traffic management systems monitor the traffic surveillance camera by using manual perception of the captured footage.
Since most intersections are equipped with surveillance cameras automatic detection of traffic accidents based on computer vision technologies will mean a great deal to traffic monitoring systems.
Numerous studies have applied computer vision techniques in traffic surveillance systems \cite{ shi2018newIcpr,liu2021smart,ghahremannezhad2020automatic,ghahremannezhad2020new,ghahremannezhad2020robust,shi2020statistical,ghahremannezhadreal,faruque2019vehicle,ghahremannezhad2021new,shi2021anomalous} for various tasks.
Automatic detection of traffic incidents not only saves a great deal of unnecessary manual labor, but the spontaneous feedback also helps the paramedics and emergency ambulances to dispatch in a timely fashion.
An automatic accident detection framework provides useful information for adjusting intersection signal operation and modifying intersection geometry in order to defuse severe traffic crashes.

\begin{figure*}[!t]
	\centering
	\includegraphics[width=0.9\linewidth]{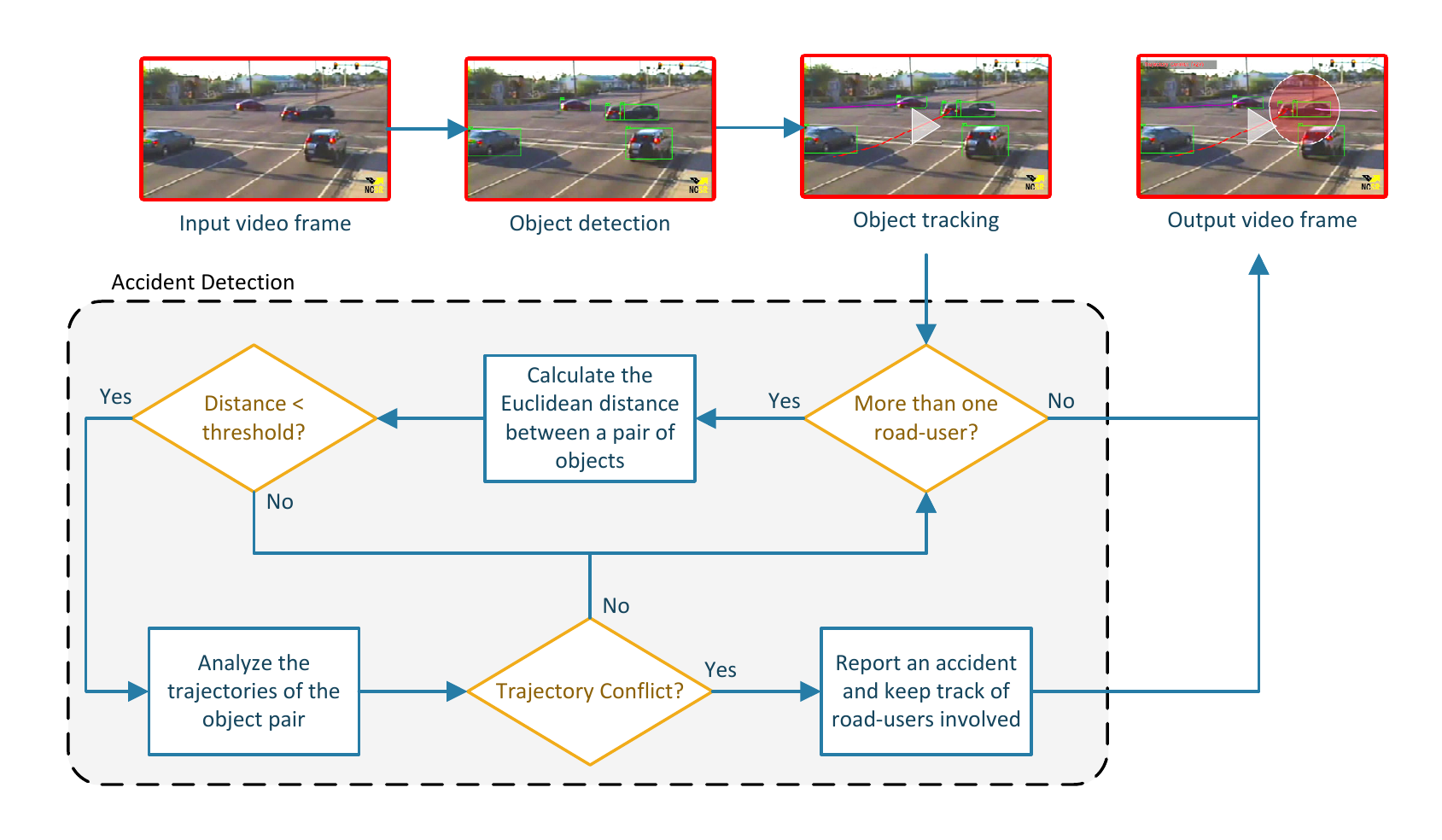}
	\caption{The system architecture of our proposed accident detection framework.}
	\label{fig_flowchart}
\end{figure*}

Considering the applicability of our method in real-time edge-computing systems, we apply the efficient and accurate YOLOv4 \cite{bochkovskiy2020yolov4} method for object detection.
The second step is to track the movements of all interesting objects that are present in the scene to monitor their motion patterns.
A new set of dissimilarity measures are designed and used by the Hungarian algorithm \cite{kuhn1955hungarian} for object association coupled with the Kalman filter approach \cite{kalman1960new} for smoothing the trajectories and predicting missed objects.
The third step in the framework involves motion analysis and applying heuristics to detect different types of trajectory conflicts that can lead to accidents.
The moving direction and speed of road-user pairs that are close to each other are examined based on their trajectories in order to detect anomalies that can cause them to crash.
\Cref{fig_flowchart} illustrates the system architecture of our proposed accident detection framework.

The layout of this paper is as follows.
In \cref{sec_met}, the major steps of the proposed accident detection framework, including object detection (\cref{ssec_det}), object tracking (\cref{ssec_trc}), and accident detection (\cref{ssec_acc}) are discussed.
\Cref{sec_exp} provides details about the collected dataset and experimental results and the paper is concluded in section \cref{sec_con}.


\section{Methodology}\label{sec_met}
This section provides details about the three major steps in the proposed accident detection framework.
These steps involve detecting interesting road-users by applying the state-of-the-art YOLOv4 \cite{bochkovskiy2020yolov4} method with a pre-trained model based on deep convolutional neural networks, tracking the movements of the detected road-users using the Kalman filter approach, and monitoring their trajectories to analyze their motion behaviors and detect hazardous abnormalities that can lead to mild or severe crashes.
The proposed framework is purposely designed with efficient algorithms in order to be applicable in real-time traffic monitoring systems.

\subsection{Road-User Detection}\label{ssec_det}
As in most image and video analytics systems the first step is to locate the objects of interest in the scene.
Since here we are also interested in the category of the objects, we employ a state-of-the-art object detection method, namely YOLOv4 \cite{bochkovskiy2020yolov4}, to locate and classify the road-users at each video frame.
The family of YOLO-based deep learning methods demonstrates the best compromise between efficiency and performance among object detectors.

\begin{figure*} [!t]
	\centering
	\includegraphics[width=0.7\linewidth]{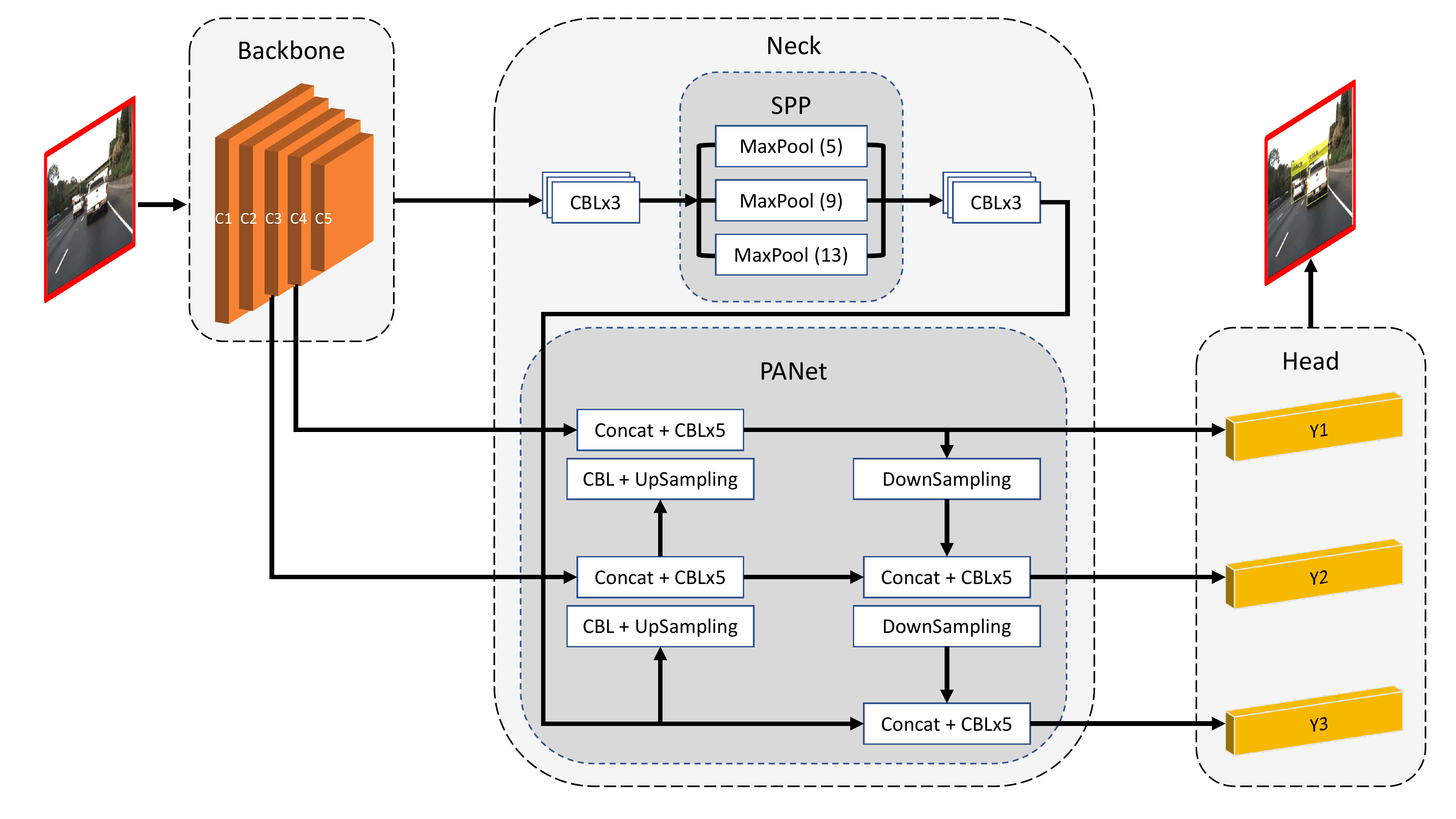}
	\caption{Architecture of the YOLOv4 model with three major component.}
	\label{fig:yolov4Arch}
\end{figure*}

The first version of the You Only Look Once (YOLO) deep learning method was introduced in 2015 \cite{redmon2016you}. 
The main idea of this method is to divide the input image into an $S \times S$ grid where each grid cell is either considered as background or used for the detecting an object.
A predefined number ($B$) of bounding boxes and their corresponding confidence scores are generated for each cell.
The intersection over union (IOU) of the ground truth and the predicted boxes is multiplied by the probability of each object to compute the confidence scores.

In later versions of YOLO \cite{redmon2017yolo9000,redmon2018yolov3} multiple modifications have been made in order to improve the detection performance while decreasing the computational complexity of the method.
Although there are online implementations such as YOLOX \cite{ge2021yolox}, the latest official version of the YOLO family is YOLOv4 \cite{bochkovskiy2020yolov4}, which improves upon the performance of the previous methods in terms of speed and mean average precision (mAP).
As illustrated in \cref{fig:yolov4Arch}, the architecture of this version of YOLO is constructed with a CSPDarknet53 model as backbone network for feature extraction followed by a neck and a head part.
The neck refers to the path aggregation network (PANet) and spatial attention module and the head is the dense prediction block used for bounding box localization and classification.
This architecture is further enhanced by additional techniques referred to as bag of freebies and bag of specials.

Here, we have applied the YOLOv4 \cite{bochkovskiy2020yolov4} model pre-trained on the MS COCO dataset \cite{lin2014microsoft} for the task of object detection.
Although the model is pre-trained on a dataset with different visual characteristics in terms of object sizes and viewing angles, YOLOv4 proved to generalize well to images with overhead perspective.
We are interested in trajectory conflicts among most common road-users at regular urban intersections, namely, vehicles, pedestrians, and cyclists.

\subsection{Road-User Tracking}\label{ssec_trc}
Multiple object tracking (MOT) has been intensively studies over the past decades \cite{luo2021multiple} due to its importance in video analytics applications.
Here we employ a simple but effective tracking strategy similar to that of the Simple Online and Realtime Tracking (SORT) approach \cite{bewley2016simple}.
The Hungarian algorithm \cite{kuhn1955hungarian} is used to associate the detected bounding boxes from frame to frame.
Additionally, the Kalman filter approach \cite{kalman1960new} is used as the estimation model to predict future locations of each detected object based on their current location for better association, smoothing trajectories, and predict missed tracks.

The inter-frame displacement of each detected object is estimated by a linear velocity model.
The state of each target in the Kalman filter tracking approach is presented as follows:
\begin{equation}\label{eq:state}
	o_i^t=\left[x_i,y_i,s_i,r_i,\dot{x_i},\dot{y_i},\dot{s_i}\right]
\end{equation}
where $x_i$ and $y_i$ represent the horizontal and vertical locations of the bounding box center, $s_i$, and $r_i$ represent the bounding box scale and aspect ratio, and $\dot{x_i},\dot{y_i},\dot{s_i}$ are the velocities in each parameter $x_i,y_i,s_i$ of object $o_i$ at frame $t$, respectively.
The velocity components are updated when a detection is associated to a target.
Otherwise, in case of no association, the state is predicted based on the linear velocity model.

Considering two adjacent video frames $t$ and $t+1$, we will have two sets of objects detected at each frame as follows:
\begin{equation}\label{eq:sets}
	\begin{gathered}
		O^t = \{o_1^t, o_2^t,\dots, o_n^t \} \\
		O^{t+1} = \{o_1^{t+1}, o_2^{t+1},\dots, o_m^{t+1} \}
	\end{gathered}
\end{equation}
Every object $o_i$ in set $O^t$ is paired with an object $o_j$ in set $O^{t+1}$ that can minimize the cost function $C(o_i,o_j)$.
The index $i \in [N] = {1,2,\dots,N}$ denotes the objects detected at the previous frame and the index $j \in [M] = {1,2,\dots,M}$ represents the new objects detected at the current frame.

In order to efficiently solve the data association problem despite challenging scenarios, such as occlusion, false positive or false negative results from the object detection, overlapping objects, and shape changes, we design a dissimilarity cost function that employs a number of heuristic cues, including appearance, size, intersection over union (IOU), and position.
The appearance distance is calculated based on the histogram correlation between and object $o_i$ and a detection $o_j$ as follows:
\begin{equation}\label{eq:app}
	C_{i,j}^A = 1 - \frac{\sum_{b}\left(H_b(o_i)-\bar{H}(o_i)\right)\left(H_b(o_j)-\bar{H}(o_j)\right)}
	{\sqrt{\sum_{b}\left(H_b(o_i)-\bar{H}(o_i)\right)^2\sum_{b}\left(H_b(o_j)-\bar{H}(o_j)\right)^2}}
\end{equation}
where $C_{i,j}^A$ is a value between 0 and 1, $b$ is the bin index, $H_b$ is the histogram of an object in the RGB color-space, and $\bar{H}$ is computed as follows:
\begin{equation}\label{eq:h}
	\bar{H}(o_k) = \frac{1}{B}\sum_{b}H_b(o_k)
\end{equation}
in which $B$ is the total number of bins in the histogram of an object $o_k$.

The size dissimilarity is calculated based on the width and height information of the objects:
\begin{equation}\label{eq:size}
	C_{i,j}^S = \frac{1}{2}\left(\frac{|h_i - h_j|}{h_i + h_j} + \frac{|w_i - w_j|}{w_i + w_j}\right)
\end{equation}
where $w$ and $h$ denote the width and height of the object bounding box, respectively.
The more different the bounding boxes of object $o_i$ and detection $o_j$ are in size, the more $C_S^{i,j}$ approaches one.
The position dissimilarity is computed in a similar way:
\begin{equation}\label{eq:pos}
	C_{i,j}^P = \frac{1}{2}\left(\frac{|x_i - x_j|}{x_i + x_j} + \frac{|y_i - y_j|}{y_i + y_j}\right)
\end{equation}
where the value of $C_{i,j}^P$ is between 0 and 1, approaching more towards 1 when the object $o_i$ and detection $o_j$ are further.
In addition to the mentioned dissimilarity measures, we also use the IOU value to calculate the Jaccard distance as follows:
\begin{equation}\label{eq:jac}
	C_{i,j}^K = 1 - \frac{Box(o_i) \cap Box(o_j)}{Box(o_i) \cup Box(o_j)}
\end{equation}
where $Box(o_k)$ denotes the set of pixels contained in the bounding box of object $k$.

The overall dissimilarity value is calculated as a weighted sum of the four measures:
\begin{equation}\label{eq:jac}
	C_{i,j} = w_aC_{i,j}^A +  w_sC_{i,j}^S + w_pC_{i,j}^P + w_aC_{i,j}^A + w_kC_{i,j}^K
\end{equation}
in which $w_a$, $w_s$, $w_p$, and $w_k$ define the contribution of each dissimilarity value in the total cost function.
The total cost function is used by the Hungarian algorithm \cite{kuhn1955hungarian} to assign the detected objects at the current frame to the existing tracks.
If the dissimilarity between a matched detection and track is above a certain threshold ($\tau_d$), the detected object is initiated as a new track.

\begin{figure}[!t]
	\centering
	\includegraphics[width=0.9\linewidth]{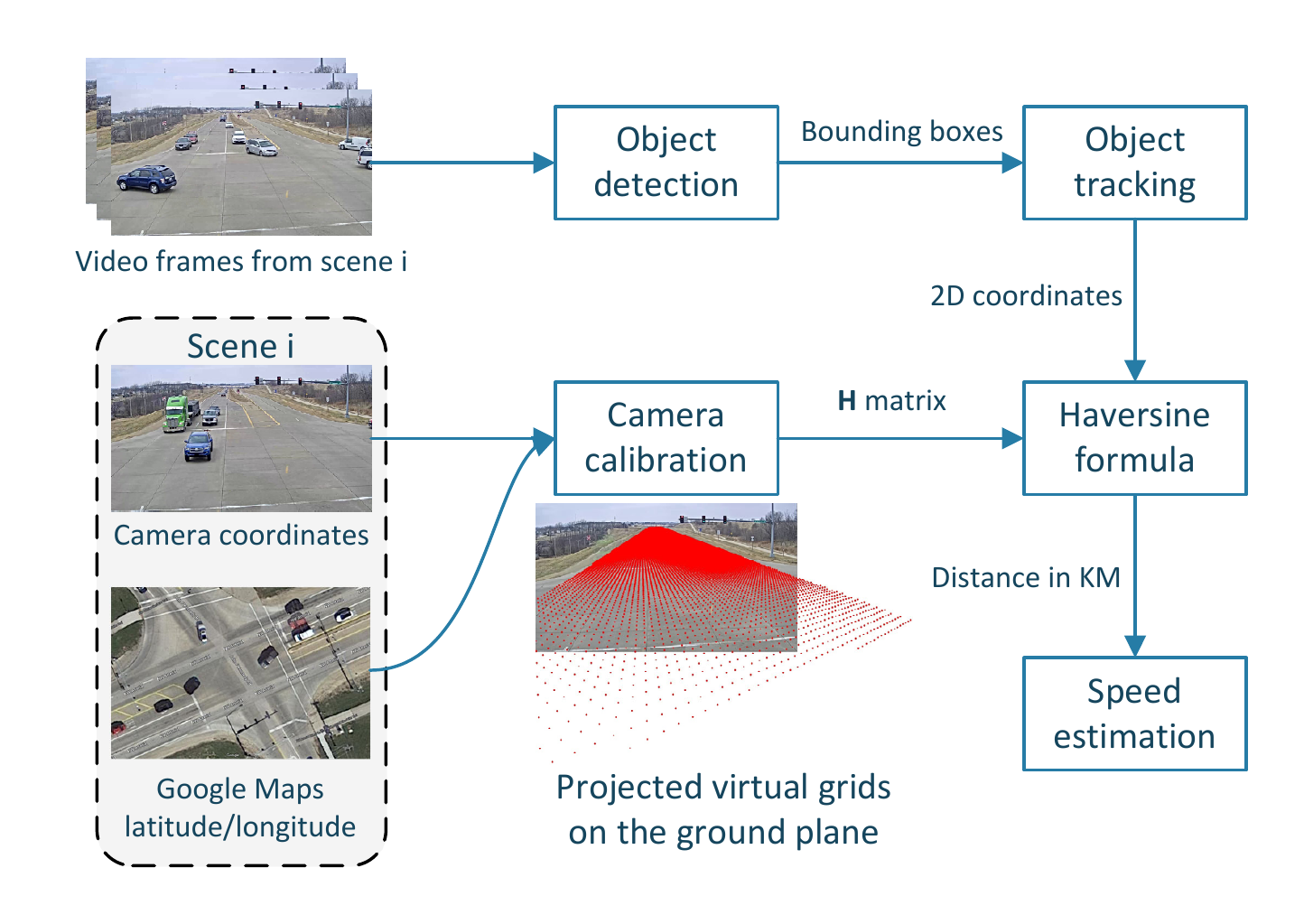}
	\caption{The workflow of the speed estimation method demonstrated on a scene from the NVIDIA AI City Challenge 2022 dataset \cite{NVIDIA}.}
	\label{fig_speed}
\end{figure}

\subsection{Accident Detection}\label{ssec_acc}
In this section, details about the heuristics used to detect conflicts between a pair of road-users are presented.
The conflicts among road-users do not always end in crashes, however, near-accident situations are also of importance to traffic management systems as they can indicate flaws associated with the signal control system and/or intersection geometry.
Logging and analyzing trajectory conflicts, including severe crashes, mild accidents and near-accident situations will help decision-makers improve the safety of the urban intersections.
The most common road-users involved in conflicts at intersections are vehicles, pedestrians, and cyclists \cite{yue2020depth}.
Therefore, for this study we focus on the motion patterns of these three major road-users to detect the time and location of trajectory conflicts.

First, the Euclidean distances among all object pairs are calculated in order to identify the objects that are closer than a threshold to each other.
These object pairs can potentially engage in a conflict and they are therefore, chosen for further analysis.
The recent motion patterns of each pair of close objects are examined in terms of speed and moving direction.

As there may be imperfections in the previous steps, especially in the object detection step, analyzing only two successive frames may lead to inaccurate results.
Therefore, a predefined number $f$ of consecutive video frames are used to estimate the speed of each road-user individually.
The average bounding box centers associated to each track at the first half and second half of the $f$ frames are computed.
The two averaged points $p$ and $q$ are transformed to the real-world coordinates using the inverse of the homography matrix $\textbf{H}^{-1}$, which is calculated during camera calibration \cite{Tang18AIC} by selecting a number of points on the frame and their corresponding locations on the Google Maps \cite{Google_2022}.
The distance in kilometers can then be calculated by applying the haversine formula \cite{gade2010non} as follows:
\begin{equation} \label{eq:haversine}
	\begin{gathered}
		h = \sin^2\left(\frac{\phi_q-\phi_p}{2}\right)+\cos\phi_p\cdot\cos\phi_q\cdot\sin^2\left(\frac{\lambda_q-\lambda_p}{2}\right)\\
		d_h(p,q) = 2r\arcsin\left(\sqrt{h}\right)	
	\end{gathered}
\end{equation}
where $\phi_p$ and $\phi_q$ are the latitudes, $\lambda_p$ and $\lambda_q$ are the longitudes of the first and second averaged points $p$ and $q$, respectively, $h$ is the haversine of the central angle between the two points, $r \approx 6371$ kilometers is the radius of earth, and $d_h(p,q)$ is the distance between the points $p$ and $q$ in real-world plane in kilometers.
The speed $s$ of the tracked vehicle can then be estimated as follows:
\begin{equation} \label{eq:speed}
	S = \frac{d_h(p,q)\times3600\times fps}{f}	
\end{equation}
where $fps$ denotes the frames read per second and $S$ is the estimated vehicle speed in kilometers per hour.
Note that if the locations of the bounding box centers among the $f$ frames do not have a sizable change (more than a threshold), the object is considered to be slow-moving or stalled and is not involved in the speed calculations.

\begin{figure*}[!t]
	\centering
	\subfloat{\includegraphics[width=1.6in]{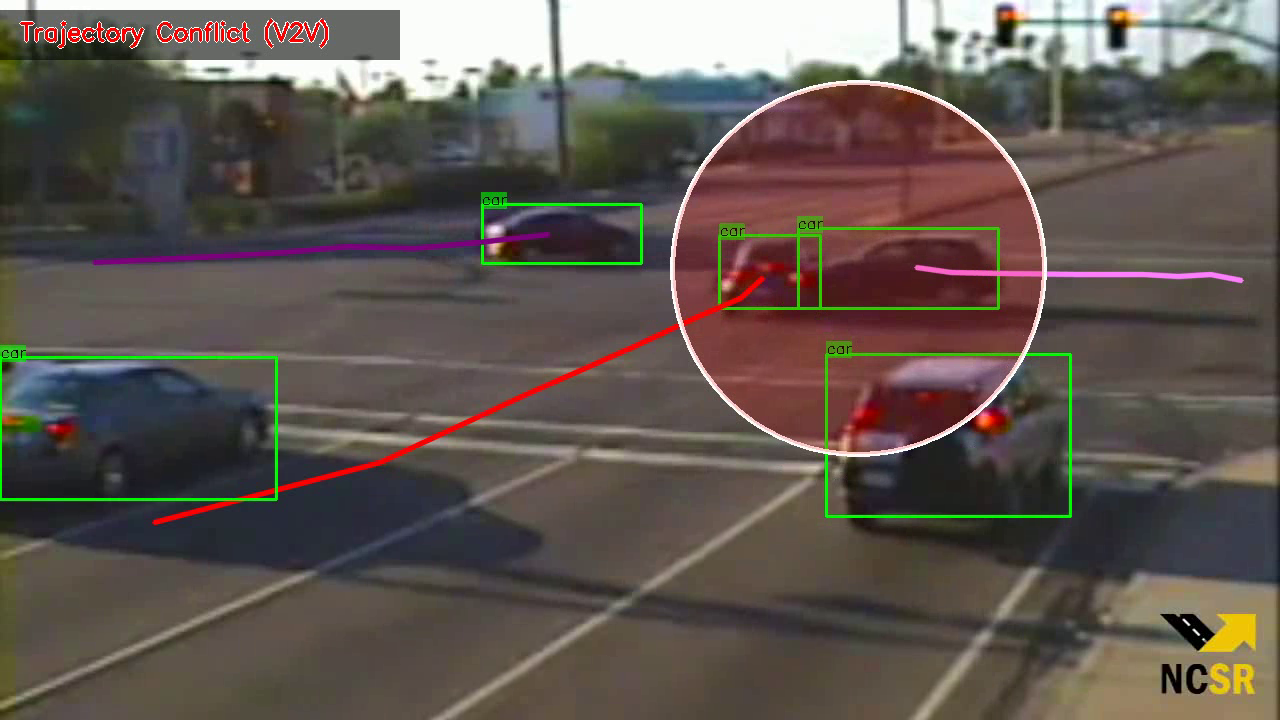}}%
	\hfil
	\subfloat{\includegraphics[width=1.6in]{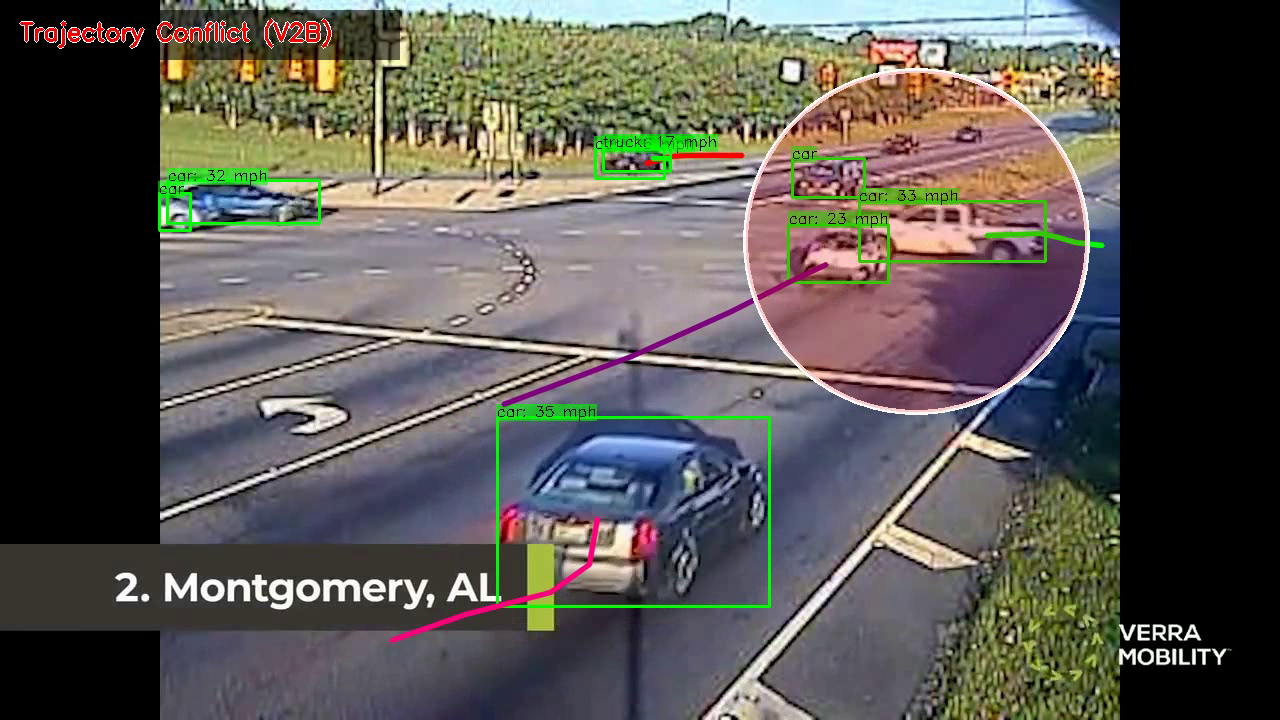}}%
	\hfil
	\subfloat{\includegraphics[width=1.6in]{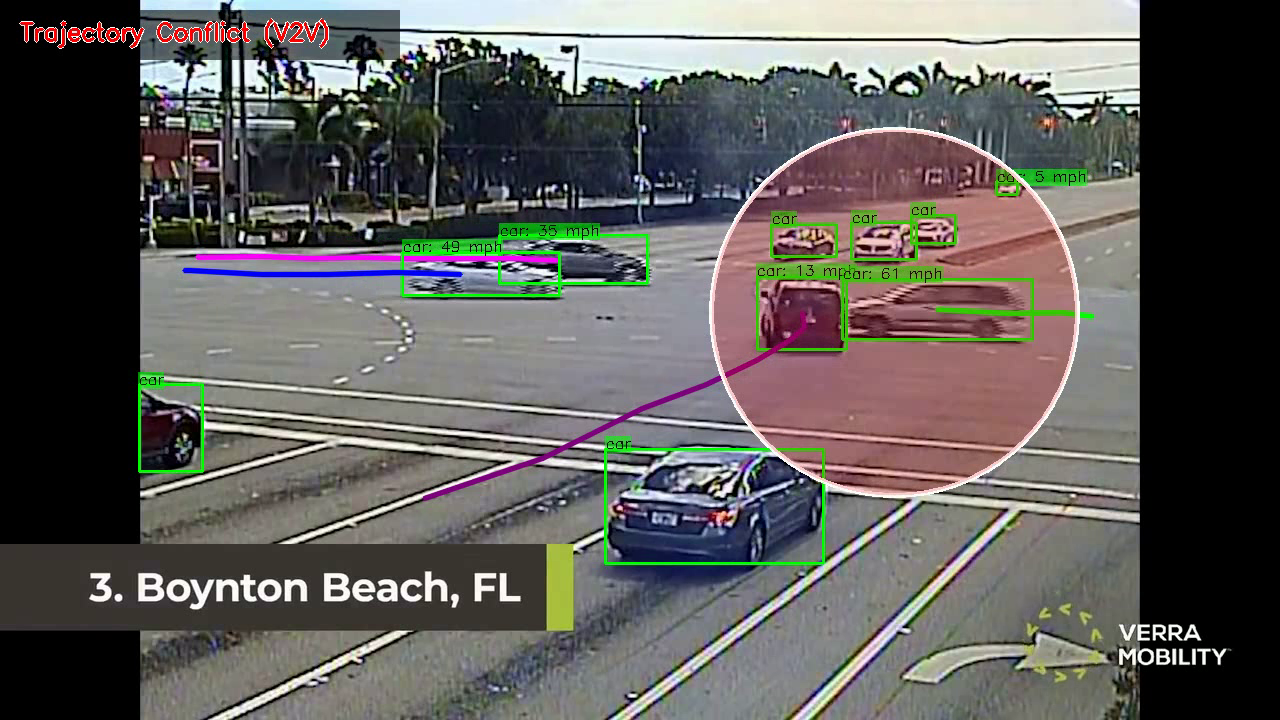}}%
	\hfil
	\subfloat{\includegraphics[width=1.6in]{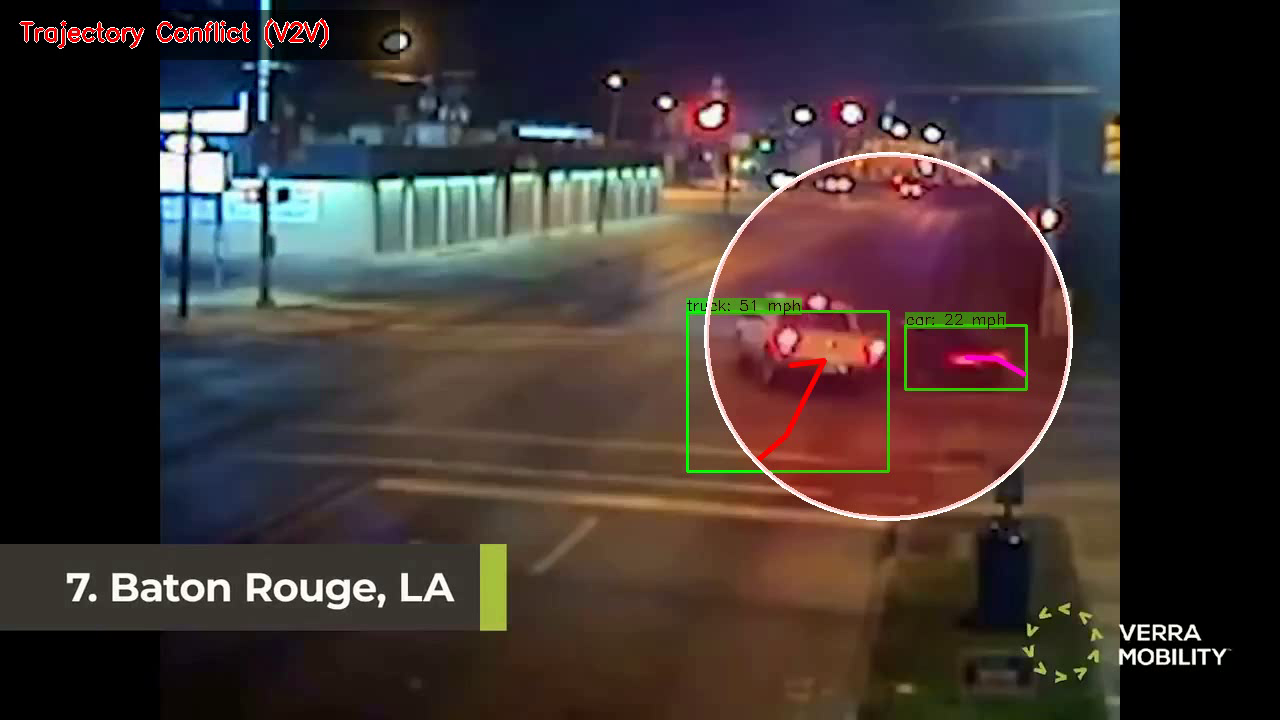}}%
	\vfil
	\subfloat{\includegraphics[width=1.6in]{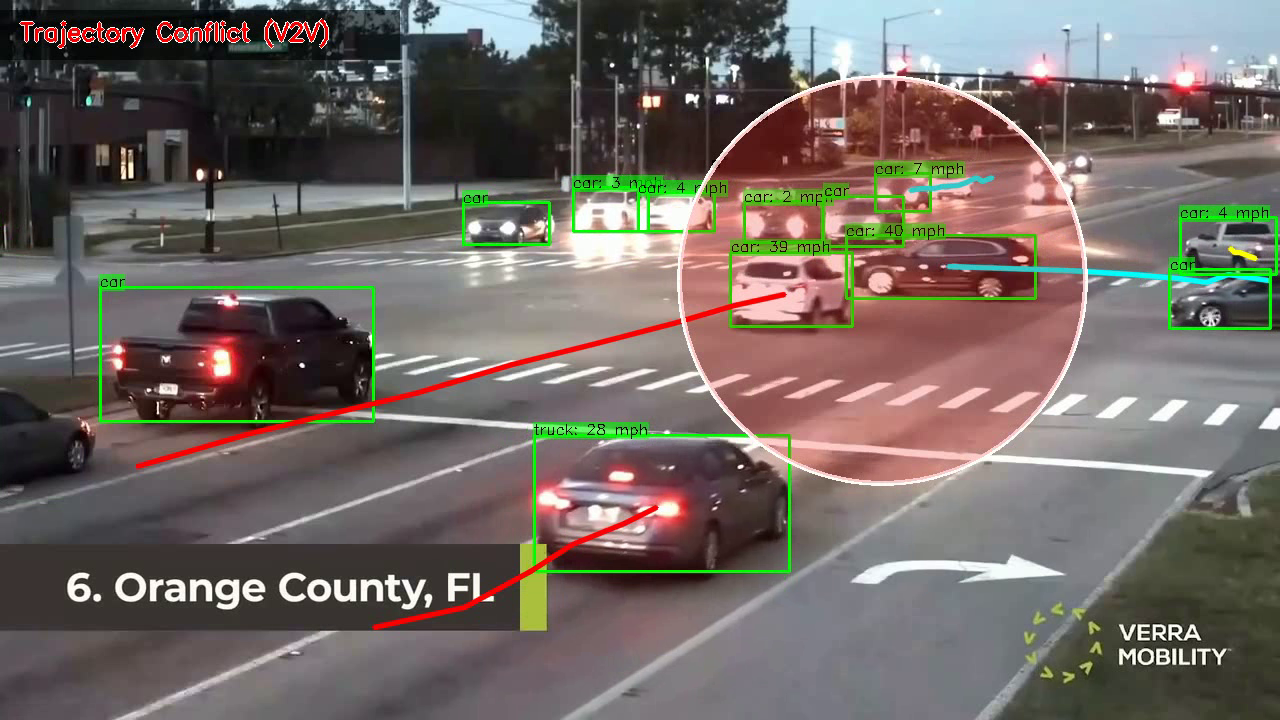}}%
	\hfil
	\subfloat{\includegraphics[width=1.6in]{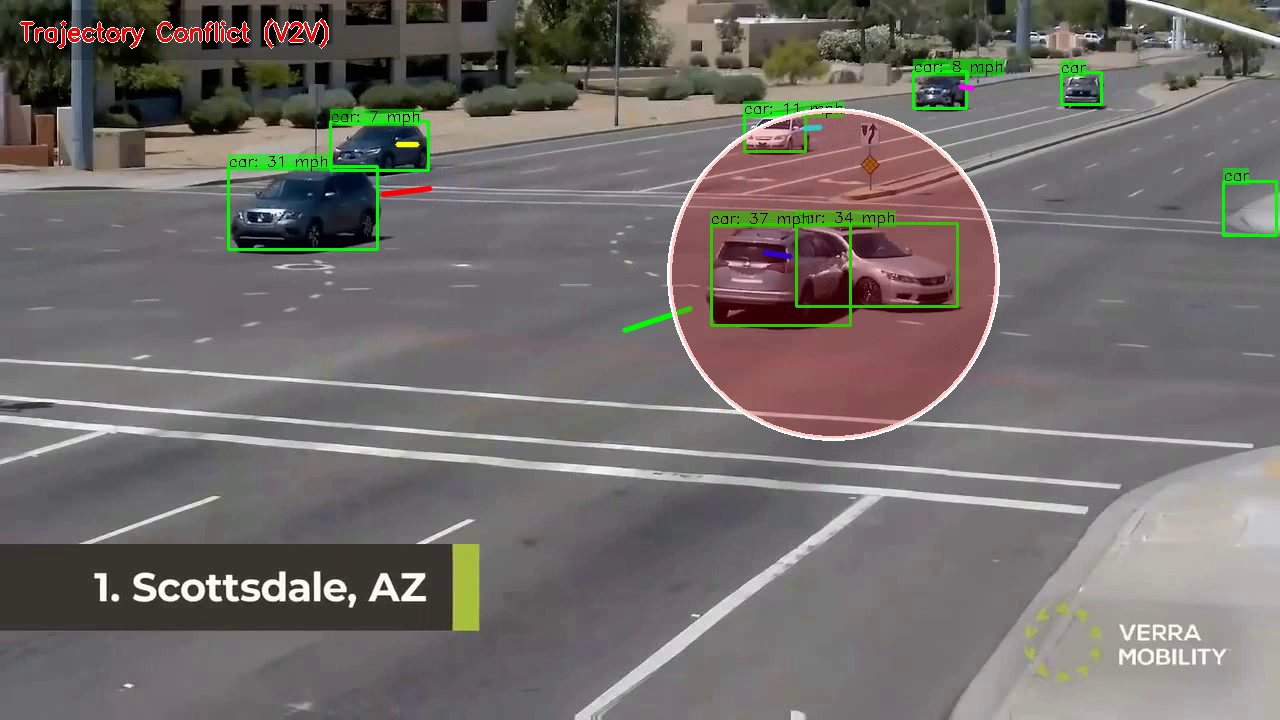}}%
	\hfil
	\subfloat{\includegraphics[width=1.6in]{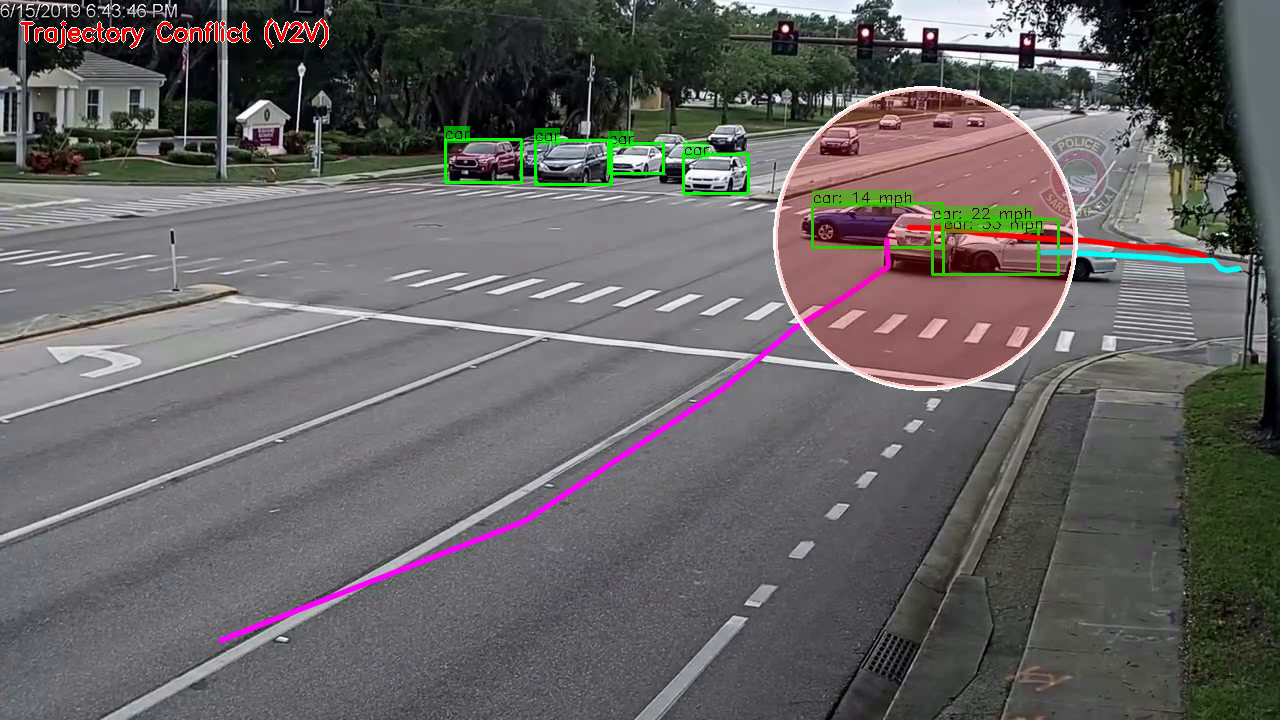}}%
	\hfil
	\subfloat{\includegraphics[width=1.6in]{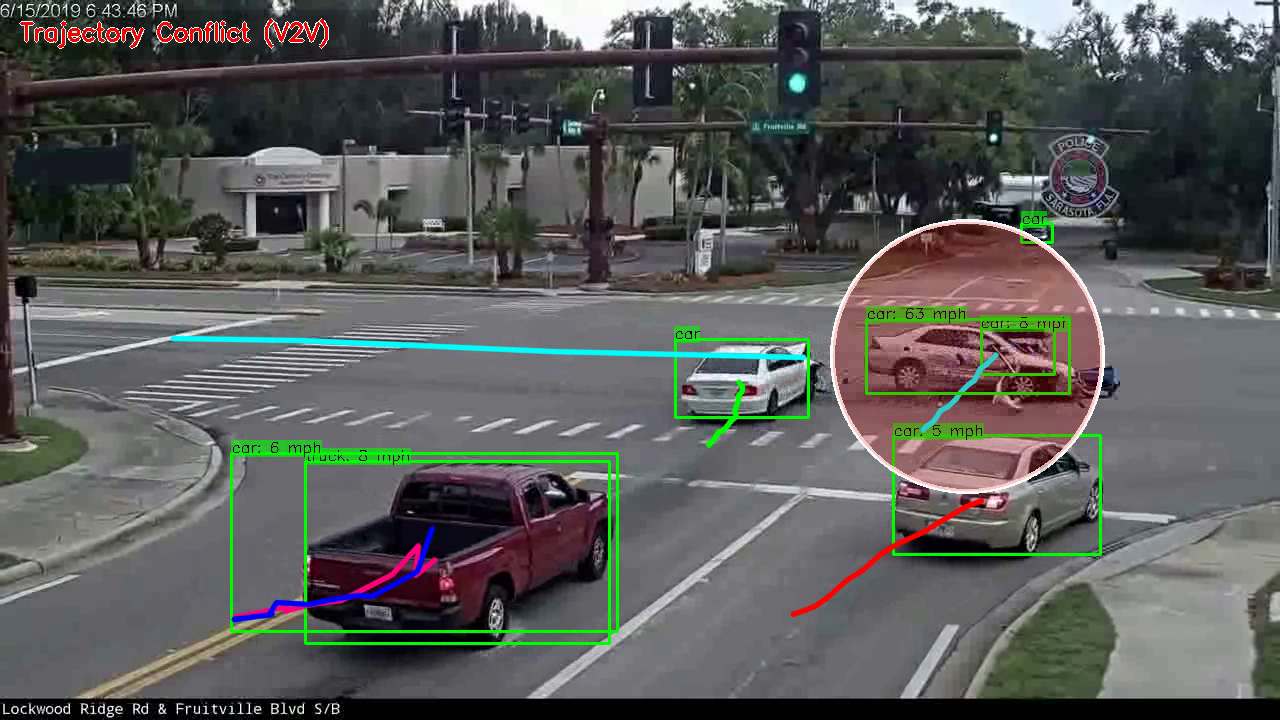}}%
	\vfil
	\subfloat{\includegraphics[width=1.6in]{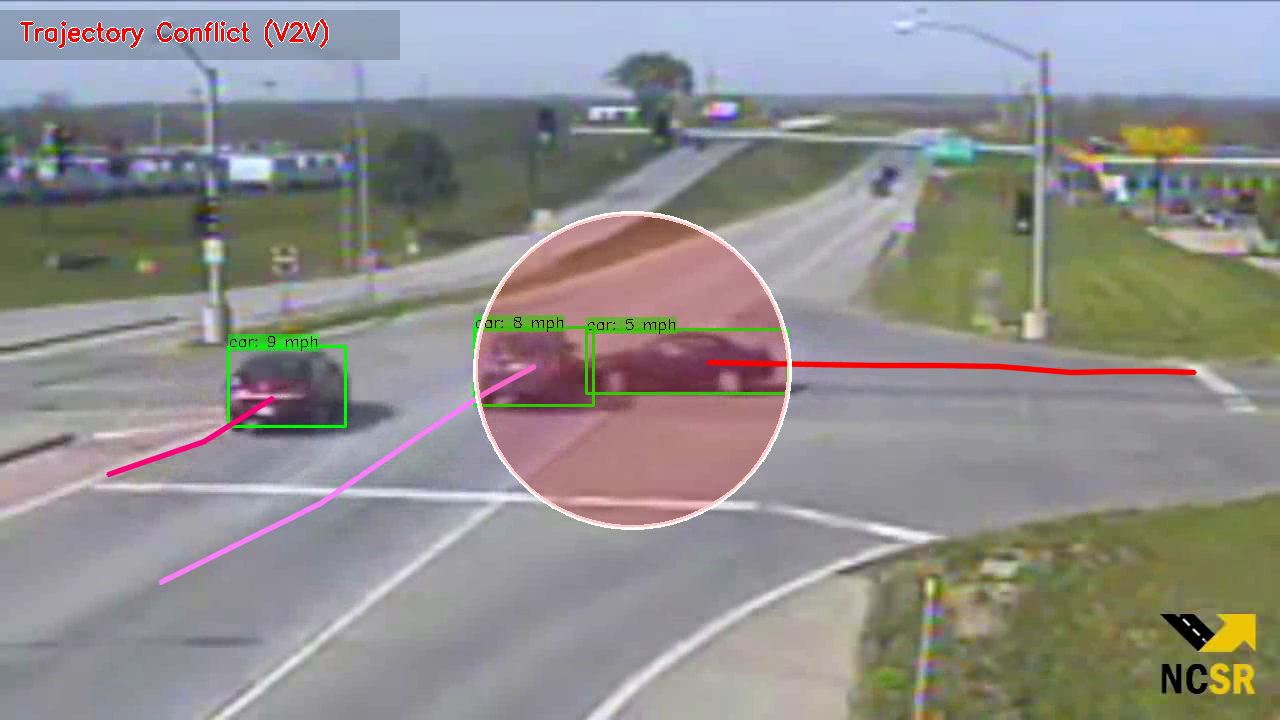}}%
	\hfil
	\subfloat{\includegraphics[width=1.6in]{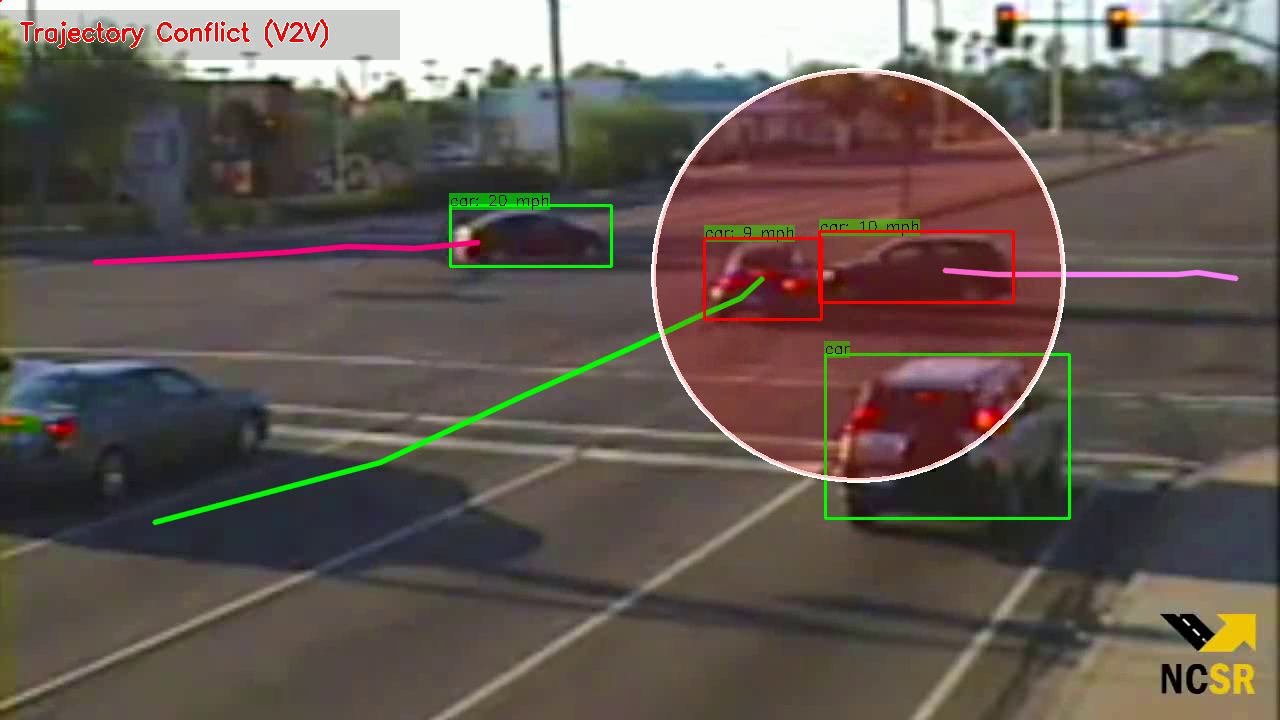}}%
	\hfil
	\subfloat{\includegraphics[width=1.6in]{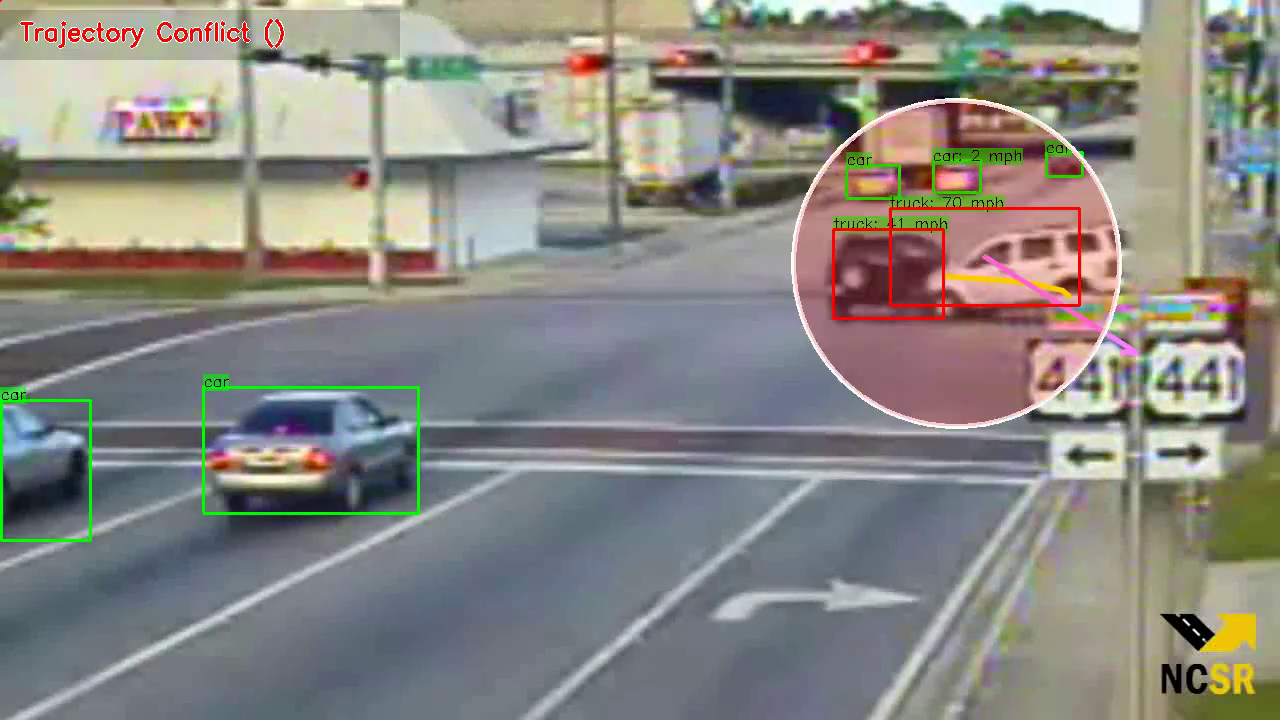}}%
	\hfil
	\subfloat{\includegraphics[width=1.6in]{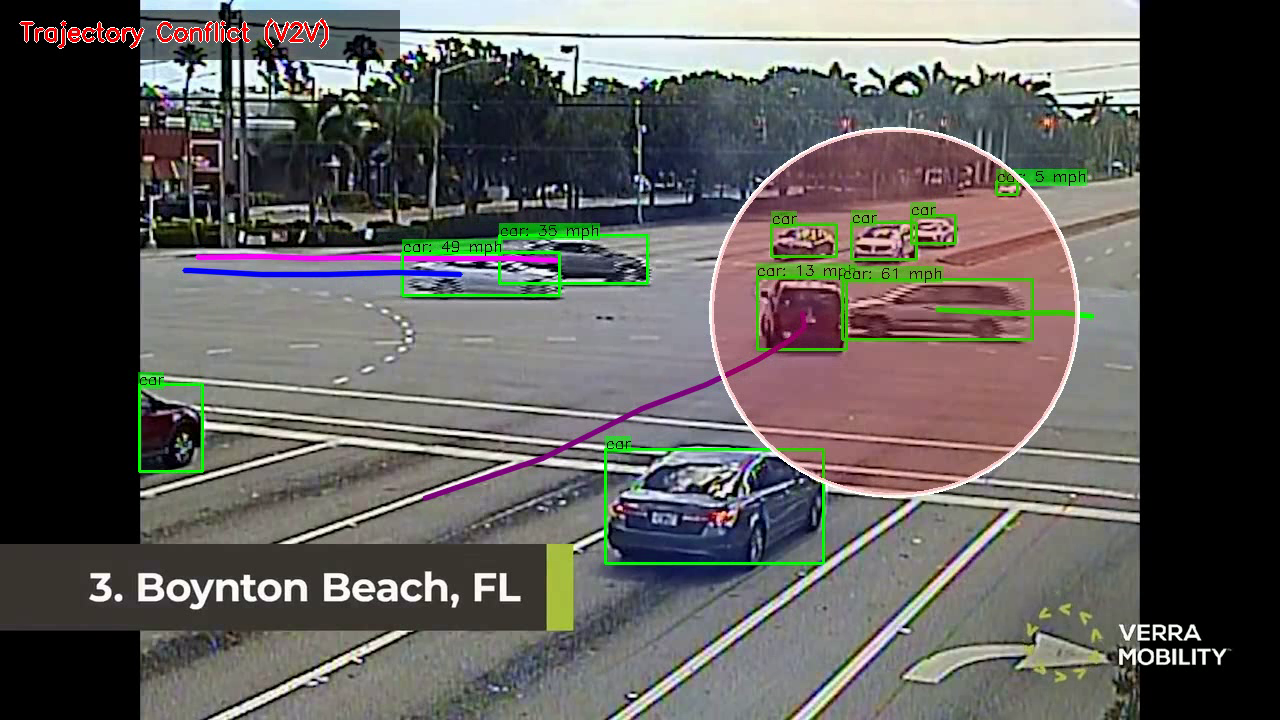}}%
	\caption{Vehicle-to-Vehicle (V2V) traffic accidents at intersections detected by our proposed framework. The red circles indicate the location of the incidents.}
	\label{fig_v2v}
\end{figure*}
 
Another factor to account for in the detection of accidents and near-accidents is the angle of collision.
Traffic accidents include different scenarios, such as rear-end, side-impact, single-car, vehicle rollovers, or head-on collisions, each of which contain specific characteristics and motion patterns.
Accordingly, our focus is on the side-impact collisions at the intersection area where two or more road-users collide at a considerable angle.
The bounding box centers of each road-user are extracted at two points: (i) when they are first observed and (ii) at the time of conflict with another road-user.
Then the approaching angle of the a pair of road-users $a$ and $b$ is calculated as follows:
\begin{equation} \label{eq:dir}
	\begin{gathered}
		m_a = \frac{\left(y_a^t - y_a^{t'}\right)}{\left(x_a^t - x_a^{t'}\right)}\\
		m_b = \frac{\left(y_b^t - y_b^{t''}\right)}{\left(x_b^t - x_b^{t''}\right)}\\
		\theta = arctan\left(\frac{m_a - m_b}{1 + m_a m_b}\right)
	\end{gathered}
\end{equation}
where $\theta$ denotes the estimated approaching angle, $m_a$ and $m_b$ are the the general moving slopes of the road-users $a$ and $b$ with respect to the origin of the video frame, $x_a^t$, $y_a^t$, $x_b^t$, $y_b^t$ represent the center coordinates of the road-users $a$ and $b$ at the current frame, $x_a^{t'}$ and $y_a^{t'}$ are the center coordinates of object $a$ when first observed, $x_b^{t''}$ and $y_b^{t''}$ are the center coordinates of object $b$ when first observed, respectively.

If the bounding boxes of the object pair overlap each other or are closer than a threshold the two objects are considered to be close.
The trajectories of each pair of close road-users are analyzed with the purpose of detecting possible anomalies that can lead to accidents.
The variations in the calculated magnitudes of the velocity vectors of each approaching pair of objects that have met the distance and angle conditions are analyzed to check for the signs that indicate anomalies in the speed and acceleration.
If the pair of approaching road-users move at a substantial speed towards the point of trajectory intersection during the previous $f$ frames and the speed of one or both shows a sudden drop at the most recent frames, a trajectory conflict is reported.
Trajectory conflicts involve near-accident and accident occurrences and include three types, namely, vehicle-to-vehicle (V2V), vehicle-to-pedestrian (V2P), and vehicle-to-bicycle (V2B).


\section{Experiments}\label{sec_exp}
Due to the lack of a publicly available benchmark for traffic accidents at urban intersections, we collected 29 short videos from YouTube that contain 24 vehicle-to-vehicle (V2V), 2 vehicle-to-bicycle (V2B), and 3 vehicle-to-pedestrian (V2P) trajectory conflict cases.
The dataset includes day-time and night-time videos of various challenging weather and illumination conditions.
Each video clip includes a few seconds before and after a trajectory conflict.
The spatial resolution of the videos used in our experiments is  $1280 \times 720$ pixels with a frame-rate of 30 frames per seconds.
We used a desktop with a 3.4 GHz processor, 16 GB RAM, and an Nvidia GTX-745 GPU, to implement our proposed method.
The average processing speed is 35 frames per second (fps) which is feasible for real-time applications.

\begin{figure}[!t]
	\centering
	\setcounter{subfigure}{0} 
	\subfloat[]{\includegraphics[width=1.6in]{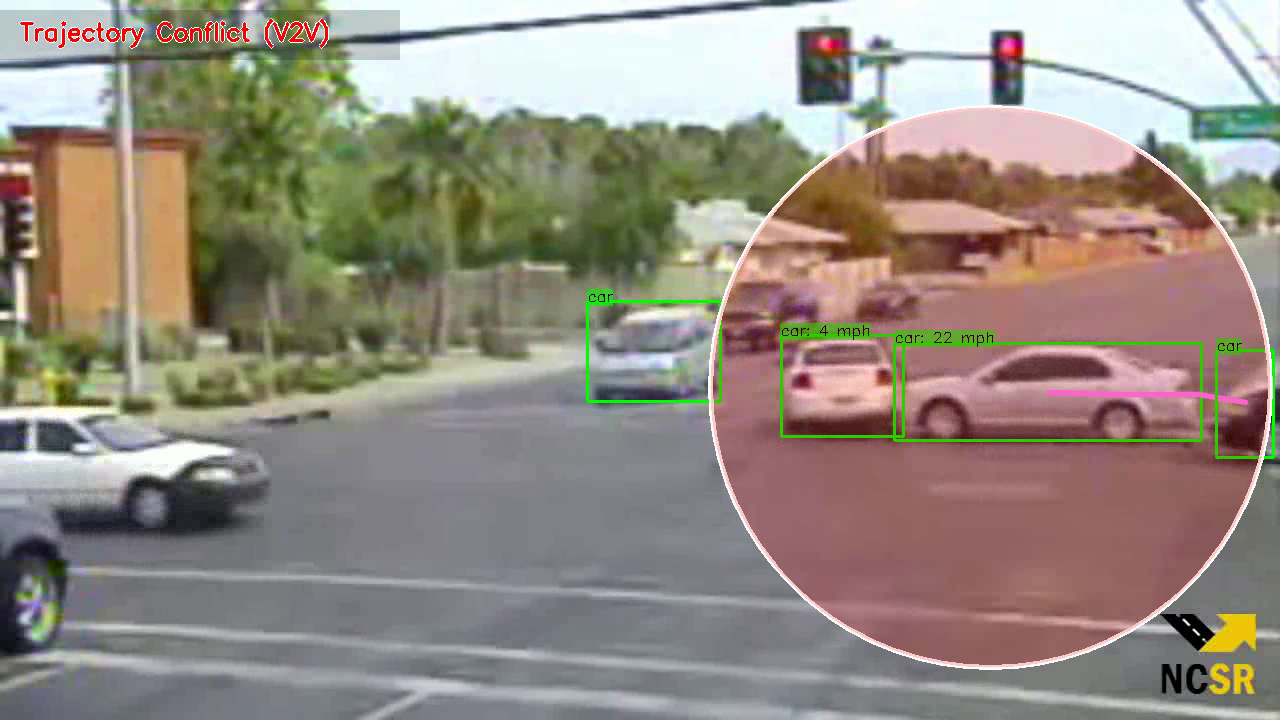}%
		\label{fig_n1}}
	\hfil
	\subfloat[]{\includegraphics[width=1.6in]{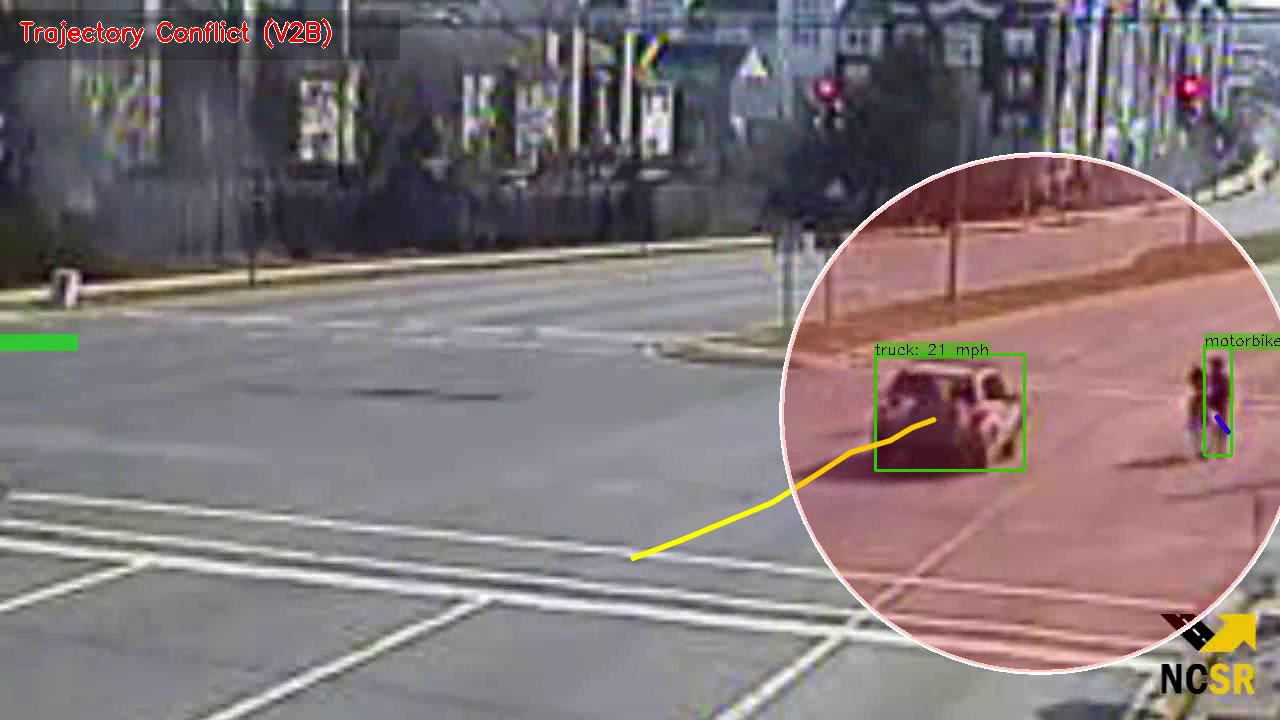}%
		\label{fig_b}}
	\vfil
	\subfloat[]{\includegraphics[width=1.6in]{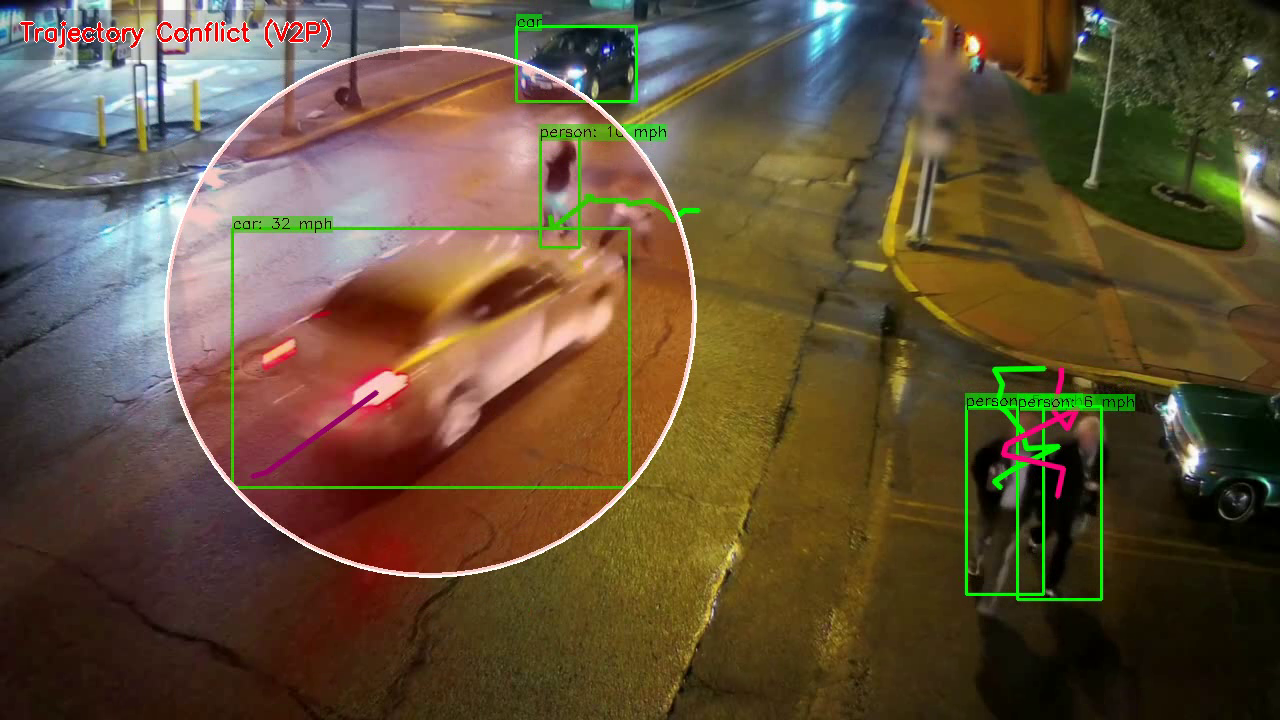}%
		\label{fig_p1}}
	\hfil
	\subfloat[]{\includegraphics[width=1.6in]{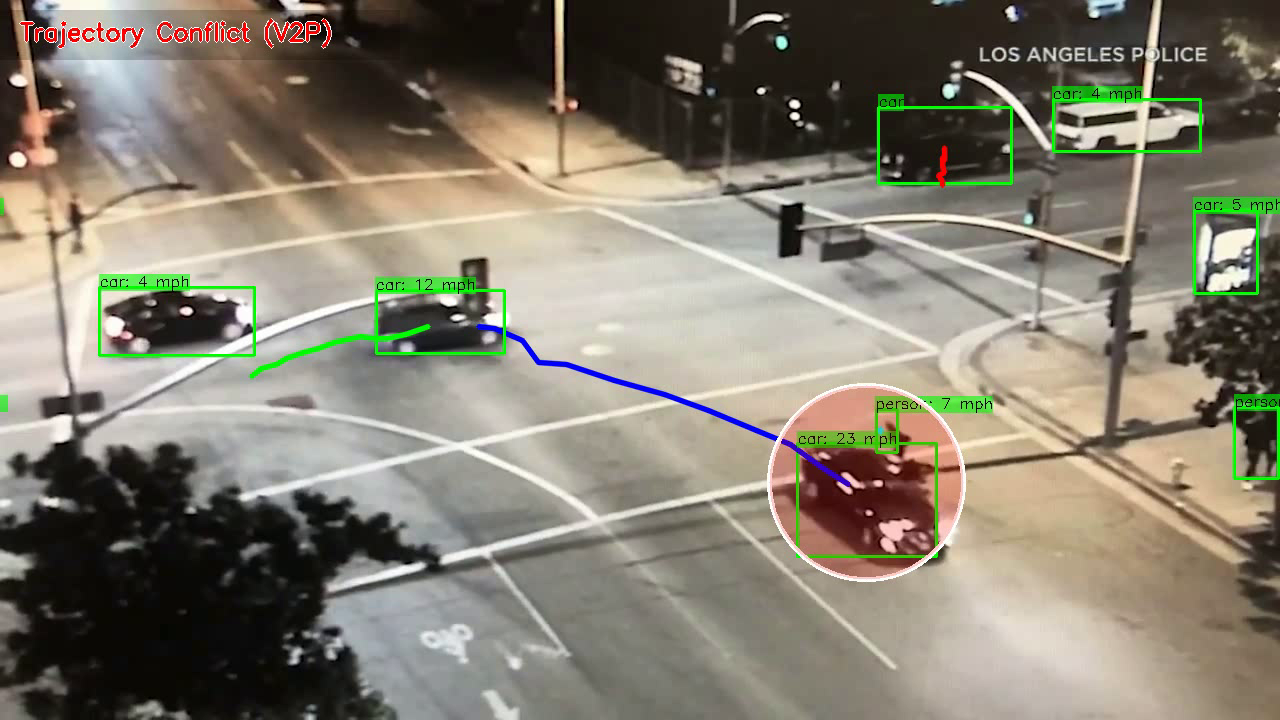}%
		\label{fig_p2}}
	\caption{\setcounter{subfigure}{0}Different types of conflicts detected at the intersections.
		\protect\subref{fig_n1} Vehicle-to-Vehicle (V2V) near-accident,
		\protect\subref{fig_b} Vehicle-to-Bicycle (V2B) near-accident,
		\protect\subref{fig_p1} and \protect\subref{fig_p2} Vehicle-to-Pedestrian (V2P) accident.
	}
	\label{fig_other}
\end{figure}

The results are evaluated by calculating Detection and False Alarm Rates as metrics:
\begin{equation} \label{eq:eval}
	\begin{gathered}
		DR = \frac{\textnormal{detected conflict cases}}{\textnormal{total number of conflicts}}\\
		FAR = \frac{\textnormal{number of false alarms}}{\textnormal{total number of conflicts}}
	\end{gathered}
\end{equation}
The proposed framework achieved a Detection Rate of $93.10\%$ and a False Alarm Rate of $6.89\%$.
The performance is compared to other representative methods in \cref{tab:perf}.
The object detection and object tracking modules are implemented asynchronously to speed up the calculations.
The trajectory conflicts are detected and reported in real-time with only 2 instances of false alarms which is an acceptable rate considering the imperfections in the detection and tracking results.
Our framework is able to report the occurrence of trajectory conflicts along with the types of the road-users involved immediately.
Additionally, it keeps track of the location of the involved road-users after the conflict has happened.
\Cref{fig_v2v} shows sample accident detection results by our framework given videos containing vehicle-to-vehicle (V2V) side-impact collisions.
Furthermore, \Cref{fig_other} contains samples of other types of incidents detected by our framework, including near-accidents, vehicle-to-bicycle (V2B), and vehicle-to-pedestrian (V2P) conflicts.

\begin{table}[!t]
	\caption{Performance comparison with other representative accident detection methods.}
	\centering
	\begin{tabular}{c c c c}
		\hline
		Methods &  Num. of videos & DR \% & FAR \%\\	
		\hline
		\hline
		Ki et al. \cite{ki2007traffic} & 1 & 63 & 6\\
		Singh et al. \cite{singh2018deep} & 7 & 77.5 & 22.5\\
		Ijjina et al. \cite{ijjina2019computer} & 45 & 71 & \textbf{0.53}\\
		Wang et al. \cite{wang2020vision} & -- & 92.5 & 7.5\\
		Pawar et al. \cite{pawar2021deep} & 7 & 79 & 20.5\\
		\textbf{Proposed method} & 29 & \textbf{93.1}  & 6.89 \\
		\hline
	\end{tabular}
	
	\label{tab:perf} 
\end{table}

\section{Conclusion}\label{sec_con}
In this paper a new framework is presented for automatic detection of accidents and near-accidents at traffic intersections.
The framework integrates three major modules, including object detection based on YOLOv4 method, a tracking method based on Kalman filter and Hungarian algorithm with a new cost function, and an accident detection module to analyze the extracted trajectories for anomaly detection.
The robust tracking method accounts for challenging situations, such as occlusion, overlapping objects, and shape changes in tracking the objects of interest and recording their trajectories.
The trajectories are further analyzed to monitor the motion patterns of the detected road-users in terms of location, speed, and moving direction.
Different heuristic cues are considered in the motion analysis in order to detect anomalies that can lead to traffic accidents.
A dataset of various traffic videos containing accident or near-accident scenarios is collected to test the performance of the proposed framework against real videos.
Experimental evaluations demonstrate the feasibility of our method in real-time applications of traffic management.


\bibliographystyle{ieeetr}
\bibliography{IEEEfull}

\end{document}